\documentclass[conference]{IEEEtran}
\usepackage{cite}

\ifCLASSINFOpdf
  \usepackage[pdftex]{graphicx}
\else
  \usepackage[dvips]{graphicx}
\fi

\usepackage{placeins}

\usepackage[cmex10]{amsmath}

\usepackage{algorithm}
\usepackage[noend]{algpseudocode}

\usepackage{array}

\usepackage{booktabs} 
\usepackage{enumitem}
\usepackage{algorithm}
\usepackage[noend]{algpseudocode}
\usepackage{graphicx,calc}

\usepackage[numbers,sort]{natbib}

\usepackage{url}

\usepackage{color}
\newlength\myheight
\newlength\mydepth
\settototalheight\myheight{Xygp}
\newcommand*\inlinegraphics[1]{%
  \settototalheight\myheight{Xygp}%
  \settodepth\mydepth{Xygp}%
  \raisebox{-\mydepth}{\includegraphics[height=\myheight*2]{#1}}%
}
\usepackage{multirow}
\usepackage{pbox}

\usepackage{url}

\usepackage[font=footnotesize]{caption}

\newcommand{\ignore}[1]{}

\begin{document}
\title{Learning Macromanagement in StarCraft\\ from Replays using Deep Learning}

\author{
\IEEEauthorblockN{Niels Justesen}
\IEEEauthorblockA{IT University of Copenhagen\\
Copenhagen, Denmark\\
noju@itu.dk}
\and
\IEEEauthorblockN{Sebastian Risi}
\IEEEauthorblockA{
IT University of Copenhagen\\
Copenhagen, Denmark\\
sebr@itu.dk}

}

\maketitle

\begin{abstract}
The real-time strategy game StarCraft has proven to be a challenging environment for artificial intelligence techniques, and as a result, current state-of-the-art solutions consist of numerous hand-crafted modules. In this paper, we show how macromanagement decisions in StarCraft can be learned directly from game replays using deep learning. Neural networks are trained on 789,571 state-action pairs extracted from 2,005 replays of highly skilled players, achieving top-1 and top-3 error rates of 54.6\% and 22.9\% in predicting the next build action. By integrating the trained network into UAlbertaBot, an open source StarCraft bot, the system can significantly outperform the game's built-in Terran bot, and play competitively against UAlbertaBot with a fixed rush strategy. To our knowledge, this is the first time macromanagement tasks are learned directly from replays in StarCraft. While the best hand-crafted strategies are still the state-of-the-art, the deep network approach is able to express a wide range of different strategies and thus improving the network's performance further with deep reinforcement learning is an immediately promising avenue for future research. Ultimately this approach could lead to strong StarCraft bots that are less reliant on hard-coded strategies. 
\end{abstract}

\section{Introduction}
Artificial neural networks have been a promising tool in machine learning for many tasks. In the last decade, the increase in computational resources as well as several algorithmic improvements, have allowed deep neural networks with many layers to be trained on large datasets. This approach, also re-branded as \emph{deep learning}, has remarkably pushed the limits within object recognition \cite{krizhevsky2012imagenet}, speech recognition \cite{hannun2014deep}, and many other domains. Combined with reinforcement learning, these techniques have surpassed the previous state-of-the-art in playing Atari games \cite{mnih2015human}, the classic board game Go \cite{silver2016mastering} and the 3D first-person shooter Doom \cite{lample2016playing}. 

An open challenge for these methods are real-time strategy (RTS) games such as StarCraft, which are highly complex on many levels because of their enormous state and actions space with a large number of units that must be controlled in real-time. Furthermore, in contrast to games like Go, AI algorithms in StarCraft must deal with hidden information; the opponent's base is initially hidden and must be explored continuously throughout the game to know (or guess) what strategy the opponent is following. The game has been a popular environment for game AI researchers with several StarCraft AI competitions  such as the \emph{AIIDE StarCraft AI Competition}\footnote{\url{http://www.cs.mun.ca/~dchurchill/starcraftaicomp/}}, \emph{CIG StarCraft RTS AI Competition}\footnote{\url{http://cilab.sejong.ac.kr/sc_competition/}} and the \emph{Student StarCraft AI Competition}\footnote{\url{http://sscaitournament.com/}}. 

However, bots participating in these competitions rely mainly on hard-coded strategies \cite{ontanon2013survey, churchill2016starcraft} and are rarely able to adapt to the opponent during the game. They usually have a modular control architecture that divides the game into smaller task areas, relying heavily on hand-crafted modules and developer domain knowledge. Learning to play the entire game with end-to-end deep learning, as it was done for Atari games \cite{mnih2015human}, is currently an unsolved challenge and perhaps an infeasible approach. A simpler approach, which we follow in this paper, is to apply deep learning to replace a specific function in a larger AI architecture. 

More specifically, we focus on applying deep learning to macromanagement tasks in \emph{StarCraft: Brood War} in the context of deciding what to produce next. A neural network is trained to predict these decisions  based on a training set extracted from replay files (i.e.\ game logs) of highly skilled human players. The trained neural network is  combined with the existing StarCraft bot UAlbertaBot, and is responsible for deciding what unit, building, technology, or upgrade to produce next, given the current state of the game. While our approach does not achieve state-of-the-art results on its own, it is a promising first step towards self-learning methods for macromanagement in RTS games. Additionally, the approach presented here is not restricted to StarCraft and can be directly applied to other RTS games as well.

\begin{figure}[!htb]
  \includegraphics[width=\columnwidth]{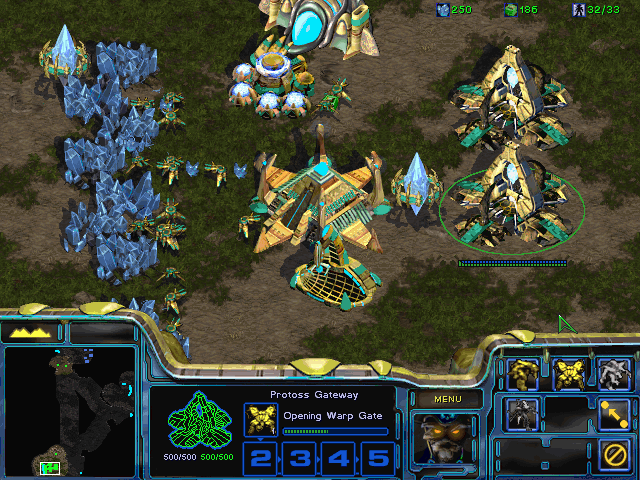} 
  \caption{A screenshot of \emph{StarCraft: Brood War}, seen from the perspective of the Protoss player. Copyright (c) Blizzard Entertainment 1998.} 
  \label{fig:starcraft}\vspace{-0.1in}
\end{figure}

\vspace{0.0in}

\section{StarCraft}

\emph{StarCraft} is a real-time strategy (RTS) game released by Blizzard in 1998. The same year an expansion set called \emph{StarCraft: Brood War} was released, which became so popular that a professional StarCraft gamer scene emerged. The game is a strategic military combat simulation in a science fiction setting. Each player controls one of three races; Terran, Protoss and Zerg. During the game, they must gather resources to expand their base and produce an army. The winner of a game is the player that manages to destroy the opponent's base. Figure \ref{fig:starcraft} shows a screenshot from a player's perspective controlling the Protoss. The screenshot shows numerous workers collecting minerals and gas resources, and some buildings used to produce combat units. To master the game, StarCraft players need quick reactions to accurately and efficiently control a large number of units in real-time. Tasks related to unit control are called \emph{micromanagement} tasks, while \emph{macromanagement} refers to the higher-level game strategy the player is following. Part of the macromanagement is the chosen \emph{build order}, i.e.\ the order in which the player produces material in the game, which can be viewed as the strategic plan a player is following. In this paper, the term \emph{build} is used to refer to any of the four types of material that can be produced: units, buildings, upgrades and technologies. Besides the opening build order, it is equally important for the player to be able to adapt to the opponent's strategy later in the game. For example, if a player becomes aware that the opponent is producing flying units it is a bad idea to exclusively produce melee troops that are restricted to  ground attacks. Players need to be able to react and adjust to the build strategies of their opponent;  learning these \emph{macromanagement} decisions is the focus of this paper. Macromanagement in StarCraft is challenging for a number of reasons, but mostly because 
areas which are not occupied by friendly units are not observable, a game mechanic known as \emph{fog-of-war}.
This restriction means that players must order units to scout the map to locate the opponent's bases. The opponent's strategy must then be deduced continuously from the partial knowledge obtained by scouting units.

Today, most StarCraft players play the sequel expansion set \emph{StarCraft II: Legacy of the Void}. While this game introduces modern 3D graphics and new units, the core gameplay is the same as in the original. For StarCraft: Brood War, bots can communicate with the game using the Brood War Application Programming Interface (BWAPI)\footnote{http://bwapi.github.io/}, which has been the foundation of several StarCraft AI competitions. 

\section{Related Work}

\subsection{Build Order Planning}
Build order planning can be viewed as a search problem, in which the goal is to find the build order that optimizes a specific heuristic. Churchill et al. applied tree search for build order planning with a goal-based approach; the search tries to minimize the time used to reach a given goal \cite{churchill2011build}. This approach is also implemented in UAlbertaBot and runs in real-time. 

Other goal-based methods that have shown promising results in optimizing opening build orders are multi-objective evolutionary algorithms \cite{kuchem2013multi,kostler2013multi,blackford2014real}. 
The downside of goal-based approaches is that goals and thus strategies are fixed, thereby preventing the bot from adapting to its opponent. Justesen et al. recently demonstrated how an approach called \emph{Continual Online Evolutionary Planning} (COEP) can continually evolve build orders during the game itself to adapt to the opponent \cite{justesen2017continual}. In contrast to a goal-based approach, COEP does not direct the search towards a fixed goal but can instead adapt to the opponent's units. 
The downside of this approach is however, that it requires a sophisticated heuristic that is difficult to design.

\subsection{Learning from StarCraft Replays}
Players have the option to save a replay file after each game in StarCraft, which enables them to watch the game without fog-of-war. Several web sites are dedicated to hosting replay files, as they are a useful resource to improve one's strategic knowledge of the game. Replay files contain the set of actions performed by both players, which the StarCraft engine can use to reproduce the exact events. Replay files are thus a great resource for machine learning if one wants to learn how players are playing the game. This section will review some previous approaches that learn from replay files.

Case-based reasoning \cite{ontanon2007case, weber2009data, hsieh2008building}, feature-expanded decision trees \cite{cho2013replay}, and several traditional machine learning algorithms \cite{cho2013replay} have been used to predict the opponent's strategy in RTS games by learning from replays. While strategy prediction is a critical part of playing StarCraft, the usefulness of applying these approaches to StarCraft bots has not been demonstrated. 

Dereszynski et al. trained Hidden Markov Models on 331 replays to learn the probabilities of the opponent's future unit productions as well as a probabilistic state transition model \cite{dereszynski2011learning}. The learned model takes as input the partial knowledge about the opponent's buildings and units and then outputs the probability that the opponent will produce a certain unit in the near future. Synnaeve et al. applied a Bayesian model for build tree prediction in StarCraft from partial observations with robust results even with 30\% noise (i.e.\ up to 30\% of the opponent's buildings are unknown) \cite{synnaeve2011bayesian}. These predictive models can be very useful for a StarCraft bot, but they do not directly determine what to produce during the game. Tactical decision making can benefit equally from combat forward models; Uriarte et al. showed how such a model can be fine-tuned using knowledge learned from replay data \cite{uriarte2015automatic}.

The approach presented in this paper addresses the complete challenge that is involved in deciding what to produce. Additionally, our approach learns a solution to this problem using deep learning, which is briefly described next. 

\subsection{Deep Learning}
Artificial neural networks are computational models loosely inspired by the functioning of biological brains. Given an input signal, an output is computed by traversing a large number of connected neural units.  The topology and connection weights of these networks can be optimized with evolutionary algorithms, which is a popular approach to evolve game-playing behaviors \cite{risi2015neuroevolution}. 
In contrast, deep learning most often refers to deep neural networks trained with gradient descent methods (e.g.\ backpropagation) on large amounts of data, which has shown remarkable success in a variety of different fields. In this case the network topologies are often hand-designed with many layers of computational units, while the parameters are learned through small iterated updates.
As computers have become more powerful and with the help of algorithmic improvements, it has become feasible to train deep neural networks to perform at a human-level in object recognition \cite{krizhevsky2012imagenet} and speech recognition \cite{hannun2014deep}. 

A combination of deep learning and reinforcement learning has achieved human-level results in Atari video games \cite{mnih2015human, mnih2016asynchronous} and beyond human-level in the classic board game Go \cite{silver2016mastering}. In the case of Go, pre-training the networks on game logs of human players to predict actions was a critical step in achieving this landmark result because it allowed further training through self-play with reinforcement learning. 

While deep learning has been successfully applied to achieve human-level results for many types of games, it is still an open question how it can be applied to StarCraft. On a much smaller scale Stanescu et al. showed how to train convolutional neural networks as game state evaluators in $\mu$RTS \cite{stanescu2016evaluating} and Usunier et al. applied reinforcement learning on small-scale StarCraft combats \cite{usunier2016episodic}. To our knowledge no prior work shows how to learn macromanagement actions from replays using deep learning.

Also worth mentioning is a technique known as imitation learning, in which a policy is trained to imitate human players. Imitation learning has been applied to Super Mario \cite{chen2017game} and Atari games \cite{bogdanovic2014deep}. These results suggest that learning to play games from human traces is a promising approach that is the foundation of the method presented in this paper.

\section{Approach}
This section describes the presented approach, which consists of two parts. First, a neural network is trained to predict human macromanagement actions, i.e.\ what to produce next in a given state. Second, the trained network is applied to an existing StarCraft bot called UAlbertaBot by replacing the module responsible for production decisions. UAlbertaBot is an open source StarCraft bot developed by David Churchill\footnote{https://github.com/davechurchill/ualbertabot} that won the annual AIIDE StarCraft AI Competition in 2013. The bot consists of numerous hierarchical modules, such as an information manager, building manager and production manager. The production manager is responsible for managing the build order queue, i.e.\ the order in which the bot produces new builds. This architecture enables us to replace the production manager with our neural network, such that whenever the bot is deciding what to produce next, the network predicts what a human player would produce. The modular design of UAlbertaBot is described in more detail in Ontan{\'o}n et al.~\cite{ontanon2013survey}.

\subsection{Dataset}
\label{sec:dataset}
This section gives an overview of the dataset used for training and how it has been created from replay files. A replay file for StarCraft contains every action performed throughout the game by each player, and the StarCraft engine can recreate the game by executing these actions in the correct order. To train a neural network to predict the macromanagement decisions made by players, state-action pairs are extracted from replay files, where a state describes the current game situation and an action corresponds to the  next build produced by the player. Additionally, states are encoded as a vector of normalized values to be processed by our neural network.

\begin{figure}
\begin{center}
  \includegraphics[width=\columnwidth]{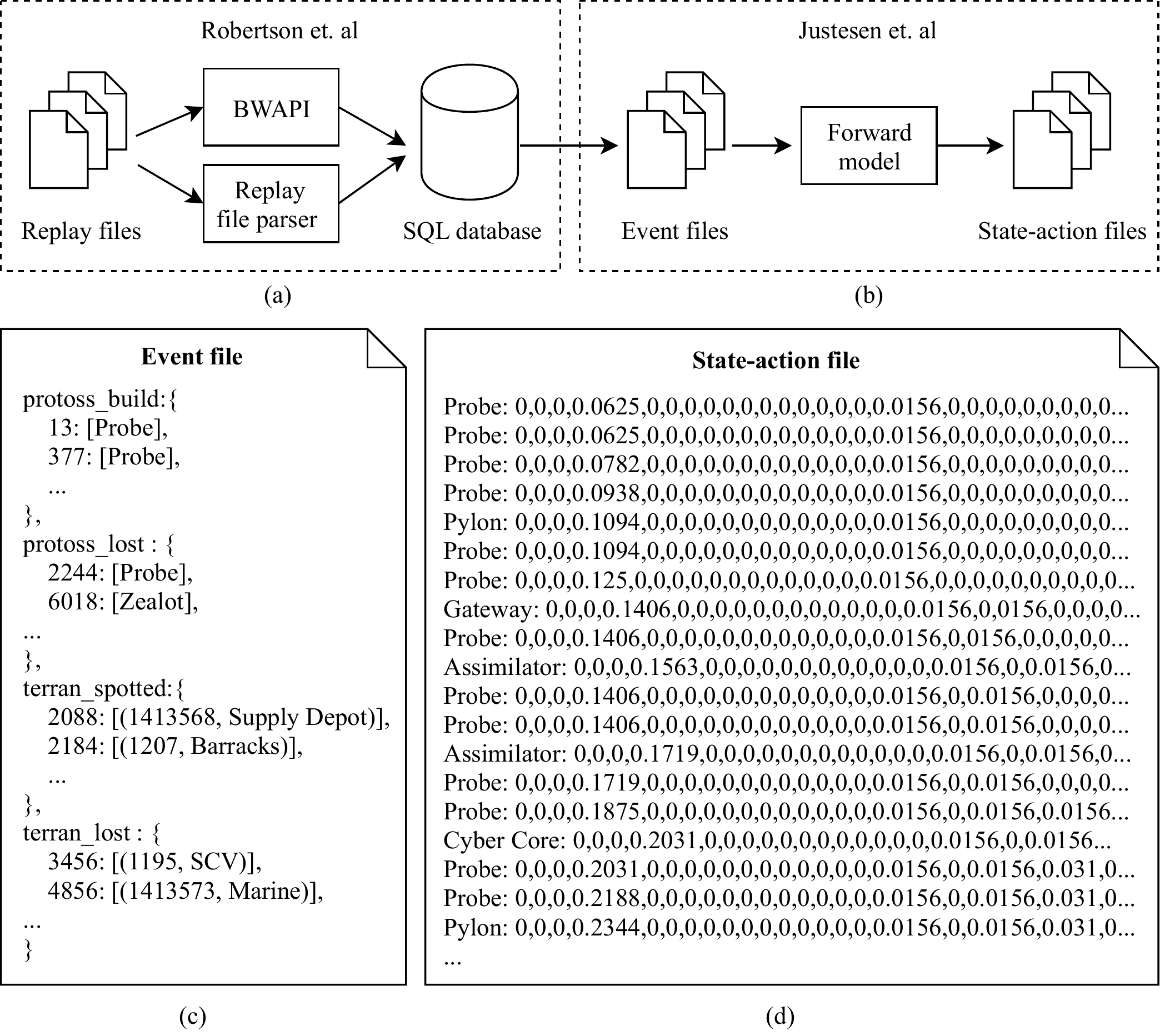} 
  \caption{An overview of the data preprocessing that converts StarCraft replays into vectorized state-action pairs. (a) shows the process of extracting data from replay files into an SQL database, which was done by Robinson et al. \cite{robertson2014improved}. (b) shows our extended data processing that first extracts events from the database into files (c) containing builds, kills and observed enemy units. All events are then run through a forward model to generate vectorized state-action pairs with normalized values (d). }\vspace{-0.00in}
  \label{fig:preprocessing}
\end{center}
\end{figure}

Replay files are in a binary format and require preprocessing before knowledge can be extracted. The dataset used in this paper is extracted from an existing dataset. Synnaeve et al. collected a repository of 7,649 replays by scraping the three StarCraft community websites GosuGamers, ICCup and TeamLiquid, which are mainly for highly skilled players including professionals \cite{synnaeve2012dataset}. A large amount of information was extracted from the repository and stored in an SQL database by Robertson et al. ~\cite{robertson2014improved}. This database contained state changes, including unit attributes, for every 24 frames in the games. Our dataset is extracted from this database, and an overview of the preprocessing steps is shown in Figure \ref{fig:preprocessing}. 

From this database, we extract all events describing material changes throughout every Protoss versus Terran game, including when (1) builds are produced by the player, (2) units and buildings are destroyed and (3) enemy units and buildings are observed. These events take the perspective of one player and thus maintain the concept of partially observable states in StarCraft. The set of events thus represent a more abstract version of the game only containing information about material changes and actions that relate to macromanagement tasks. The events are then used to simulate abstract StarCraft games via the build order forward model presented in Justesen and Risi~\cite{justesen2017continual}. Whenever the player takes an action in these abstract games, i.e.\ produces something, the action and state pair is added to our dataset. The state describes the player's own material in the game: the number of each unit, building, technology, and upgrade present and under construction, as well as enemy material observed by the player. 

The entire state vector consists of a few sub-vectors described here in order, in which the numbers represent the indexes in the vector:

\begin{enumerate}
\item \textbf{0-31:} The number of units/buildings of each type present in the game controlled by the player.
\item \textbf{32-38:} The number of each technology type researched in the game by the player. 
\item \textbf{39-57:} The number of each upgrade type researched in the game by the player. For simplicity, upgrades are treated as a one-time build and our state description thus ignores level 2 and 3 upgrades.
\item \textbf{58-115:} The number of each build in production by the player. 
\item \textbf{116-173:} The progress of each build in production by the player. If a build type is not in production it has a value of 0. If several builds of the same type are under construction, the value represents the progress of the build that will be completed first.
\item \textbf{174-206:} The number of enemy units/buildings of each type observed.
\item \textbf{207-209:} The number of supply used by the player and the maximum number of supplies available. Another value is added which is the supply left, i.e.\ the difference between supply used and maximum supplies available.
\end{enumerate}

All values are normalized into the interval [0, 1]. The preprocessed dataset contains 2,005 state-action files with a total of 789,571 state-action pairs. Six replays were excluded because the Protoss player used the rare \emph{mind control} spell on a Terran SCV that allows the Protoss player to produce Terran builds. 
While the data preprocessing required for training is a relatively long process, the same data can be gathered directly by a playing (or observing) bot during a game. 

\subsection{Network Architecture}
Since our dataset contains neither images nor sequential data, a simple multi-layered network architecture with fully-connected layers is used. Our game state contains all the material produced and observed by the player throughout the game, unless it has been destroyed, and thus there is no need for recurrent connections in our model. The network that obtained the best results has four hidden layers. The input layer has 210 units, based on the state vector described in Section \ref{sec:dataset}, which is followed by four hidden layers of 128 units with the ReLU activation function. The output layer has one output neuron for each of the 58 build types a Protoss player can produce and uses the softmax activation function. The output of the network is thus the probability of producing each build in the given state. 

\subsection{Training}
The dataset of 789,571 state-action pairs is split into a training set of 631,657 pairs (80\%) and a test set of 157,914 pairs (20\%). The training set is exclusively used for training the network, while the test set is used to evaluate the trained network. The state-action pairs, which come from 2,005 different Protoss versus Terran games, are not shuffled prior to the division of the data to avoid that actions from the same game end up in both the training and test set.

The network is trained on the training set, which is shuffled before each epoch. Xavier initialization is used for all weights in the hidden layers and biases are initialized to zero. The learning rate is 0.0001 with the Adam optimization algorithm \cite{kingma2014adam} and a batch size of 100. The optimization algorithm uses the cross entropy loss function $-\sum_i y'_i \log(y_i)$, where $y$ is the output vector of the network and $y'$ is the one-hot target vector. The problem is thus treated as a classification problem, in which the network tries to predict the next build given a game state. In contrast to classical classification problems, identical data examples (states) in our dataset can have different labels (builds), as human players execute different strategies and also make mistakes while playing. Also, there is no correct build for any state in StarCraft, but some builds are much more likely to be performed by players as they are more likely to result in a win. The network could also be trained to predict whether the player is going to win the game, but how to best incorporate this in the decision-making process is an open question. Instead here we focus on predicting actions made by human players, similarly to the supervised learning step in AlphaGo  \cite{silver2016mastering}.

\begin{figure*}
\begin{center}
  \includegraphics[width=\textwidth]{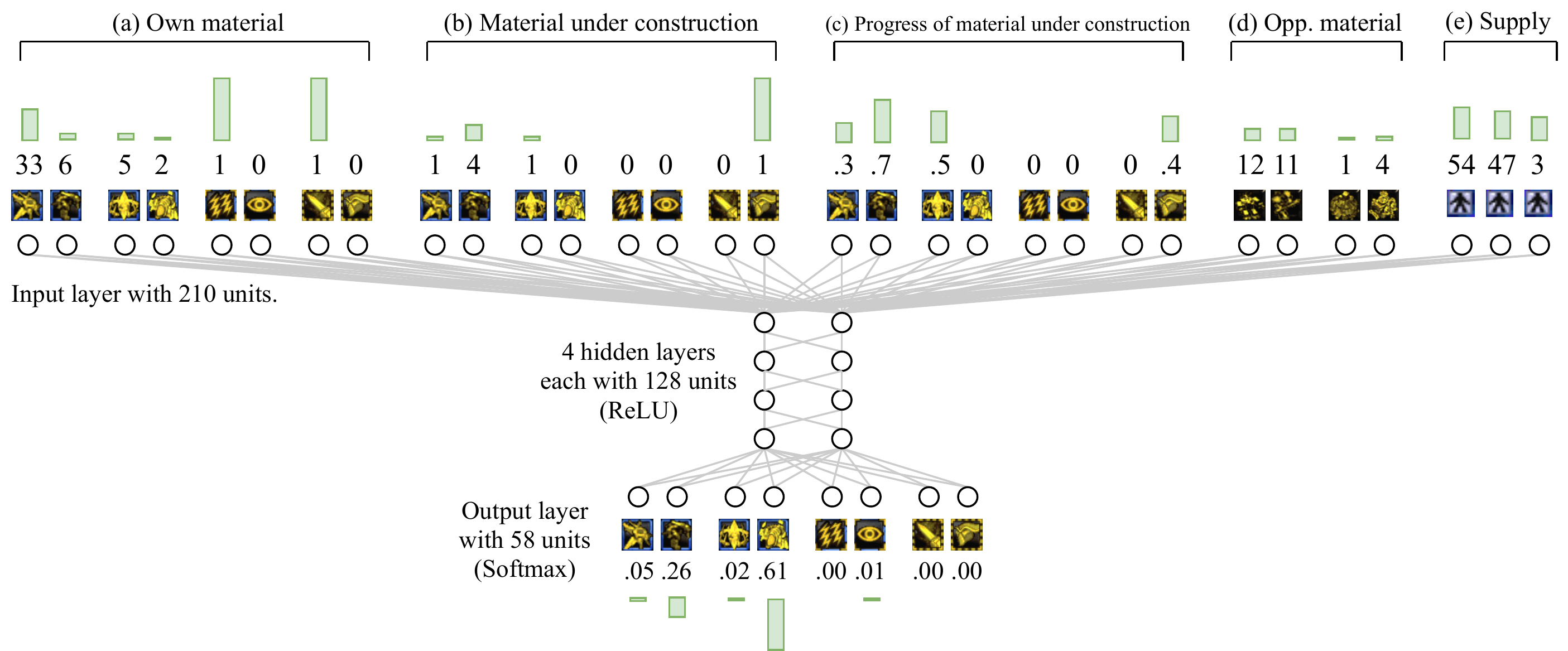} 
  \caption{Neural Network Architecture. The input layer consists of a vectorized state containing normalized values representing the number of each unit, building, technology, and upgrade in the game known to the player. Only a small subset is shown on the diagram for clarity. Three inputs also describe the player's supplies. The neural network has four hidden fully-connected layers with 128 units each using the ReLU activation function. These layers are followed by an output layer using the softmax activation function and the output of the network is the prediction of each build being produced next in the given state.  }\vspace{-0.10in}
  \label{fig:nn}
\end{center}
\end{figure*}

\subsection{Applying the Network to a StarCraft Bot}
Learning to predict actions in games played by humans is very similar to the act of learning to play. 
However, this type of imitation learning does have its limits as the agent does not learn to take optimal actions, but instead to take the most probable action (if a human was playing). However, applying the trained network as a macromanagement module of an existing bot could be an important step towards more advanced approaches. 

In this paper, we build on the UAlbertaBot, which has a
production manager that manages a queue of builds that the bots must produce in order. The production manager, which normally uses a goal-based search, is modified to use the network trained on replays  instead. The production manager in UAlbertaBot is also extended to act as a web client; whenever the module is asked for the next build, the request is forwarded, along with a description of the current game state, to a web server that feeds the game state to the neural network and then returns a build prediction to the module. Since the network is only trained on Protoss versus Terran games, it is only tested in this matchup. Our approach can however easily be applied to the other matchups as well. UAlbertaBot does not handle some of the advanced units well, so these where simply excluded from the output signals of the network. The excluded units are: archons, carriers, dark archons, high templars, reavers and shuttles. After these are excluded from the output vector, values are normalized to again sum to 1. An important  question is how to select one build action based on the network's outputs. Here two action selection policies are tested:

\textbf{Greedy action selection:} The build with the highest probability is always selected. This approach creates a deterministic behavior with a low variation in the units produced. A major issue of this approach is that rare builds such as upgrades will likely never be selected.

\textbf{Probabilistic action selection:} Builds are selected with the probabilities of the softmax output units. In the example in Figure~\ref{fig:nn}, a probe will be selected with a 5\% probability and a zealot with 26\% probability. With a low probability, this approach will also select some of the rare builds, and can express a wide range of strategies. Another interesting feature is that it is stochastic and harder to predict by the opponent.
\section{Results}

\subsection{Build Prediction}

The best network managed to reach a top-1 error rate of 54.6\% (averaged over five runs) on the test set, which means that it is able to guess the next build around half the time, and with top-3 and top-10 error rates of 22.92\% and 4.03\%. For a simple comparison, a baseline approach that always predicts the next build to be a probe, which is the most common build in the game for Protoss, has  a top-1 error rate of 73.9\% and thus performs significantly worse. Predicting randomly with uniform probabilities achieves a top-1 error rate of 98.28\%. 
Some initial experiments with different input layers show that we obtain worse error rates by omitting parts of the state vector described in \ref{sec:dataset}. For example,  when opponent material is excluding from the input layer the networks top-1 error increases to an average of 58.17\%. Similarly, omitting the material under construction (together with the progress) increases the average top-1 error rate to 58.01\%. The results are summarized in Table \ref{tab:inputs} with error rates averaged over five runs for each input layer design. The top-1, top-3 and top-10 error rates in the table show the networks' ability to predict using one, three and ten guesses respectively, determined by their output. All networks were trained for 50 epochs as the error rates stagnated prior to this point. Overfitting is minimal with a difference less than 1\% between the top-1 training and test errors. 

\begin{figure}
\begin{center}
  \includegraphics[width=\columnwidth]{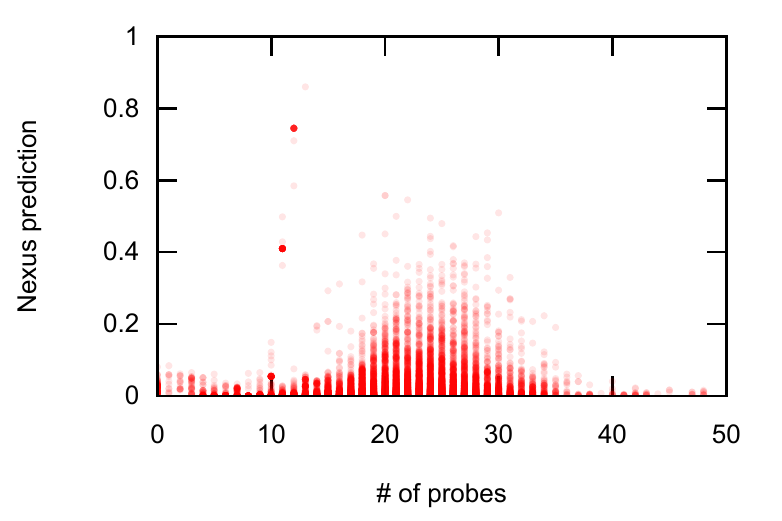} 
  \caption{The prediction of the next build being a Nexus (a base expansion) predicted by the trained neural network. Each data point corresponds to one prediction from one state. These states have only one Nexus and are taken from the test set. The small spike around 11 and 12 probes shows that the network predicts a fast expansion build order if the Protoss player has not build any gateways at this point.}\vspace{-0.2in}
  \label{fig:nexus_probes}
\end{center}
\end{figure}

To gain further insights into the policy learned by the network, the best network's prediction of building a new base given a varying number of probes is plotted in Figure~\ref{fig:nexus_probes}. States are taken from the test set in which the player has only one base. The network successfully learned that humans usually create a base expansion when they have around 20-30 probes.

\begin{table}[t]
  \centering
  
\setlength{\tabcolsep}{0.5em} 
{\renewcommand{\arraystretch}{1.0}
  \begin{tabular}{lccc}
\hline
Input & Top-1 error & Top-3 error & Top-10 error \\
\hline
\hline

\textbf{a+b+c+d+e} & \textbf{54.60\% $\pm$ 0.12\%} & \textbf{22.92\% $\pm$ 0.09\%} & \textbf{4.03\% $\pm$ 0.14\%} \\
a+b+c+e & 58.17\% $\pm$ 0.16\% & 24.92\% $\pm$ 0.10\% & 4.23\% $\pm$ 0.04\%\\
a+d & 58.01\% $\pm$ 0.42\% & 24.95\% $\pm$ 0.31\% & 4.51\% $\pm$ 0.16\%\\ 
a & 60.81\% $\pm$ 0.09\% & 26.64\% $\pm$ 0.11\% & 4.65\% $\pm$ 0.21\%\\ 
\hline
Probe & 73.90\% $\pm$ 0.00\% & 73.90\% $\pm$ 0.00\% & 73.90\% $\pm$ 0.00\% \\
Random & 98.28\% $\pm$ 0.04\% & 94.87\% $\pm$ 0.05\% & 82.73\% $\pm$ 0.08\%\\
\hline
\\
\end{tabular}
}

  \caption{ The top-1, top-3 and top-10 error rates of trained networks (averaged over five runs) with different combinations of inputs. (a) is the player's own material, (b) is material under construction, (c) is the progress of material under construction, (d) is the opponent's material and (e) is supply. The input layer is visualized in Figure~\ref{fig:nn}. \emph{Probe} is a baseline predictor that always predicts the next build to be a probe and \emph{Random} predicts randomly with uniform probabilities. The best results (in bold) are achieved by using all the input features. }\vspace{-0.0in}
\label{tab:inputs}
\end{table}

\subsection{Playing StarCraft}
UAlbertaBot is tested playing the Protoss race against the built-in Terran bot, with the trained network as production manager. Both the greedy and probabilistic actions selection strategies are tested in 100 games in the two-player map Astral Balance. The results, summarized in Table~\ref{tab:results}, demonstrates that the  probabilistic strategy is clearly superior, winning 68\% of all games. This is significant at $p \leq 0.05$ according to the two-tailed Wilcoxon Signed-Rank. The greedy approach, which always selects the action with the highest probability, does not perform as well. While the probabilistic strategy is promising, it is important to note that an UAlbertaBot playing as Protoss and following a powerful hand-designed strategy (dragoon rush), wins  100\% of all games against the built-in Terran bot. 

To further understand the difference between the two approaches, the builds selected by each selection strategy are analyzed. A subset of these builds are shown in Table~\ref{tab:builds-produced}. The probabilistic strategy clearly expresses a more varied strategy than the greedy one. Protoss players often prefer a good mix of zealots and dragoons as it creates a good dynamic army, and the greedy strategy clearly fails to achieve this. Additionally, with  the greedy approach the bot never produces any upgrades, because they are too rare in a game to ever become the most probable build. The blind probabilistic approach (which ignores knowledge about the opponent by setting these inputs to zero) reached a lower win rate of just 59\%, further corroborating  that the opponent's units and buildings are important for macromanagement decision making.
We also tested the probabilistic approach against UAlbertaBot with the original production manager configured to follow a fixed marine rush strategy, which was the best opening strategy for UAlbertaBot when playing Terran. Our approach won 45\% of 100 games, demonstrating that it can play competitively against this aggressive rush strategy, learning from human replays alone. 

\begin{table}[t]
  \centering
  \begin{tabular}{lccc}
\hline
Action selection & Built-in Terran \\
\hline
\hline
Probabilistic & 68\% \\
Probabilistic (blind) & 59\%\\
Greedy & 53\% \\
Random & 0\% \\
\hline
UAlbertaBot (dragoon rush) & 100\%
\\
\end{tabular}
  \caption{The win percentage of UAlbertaBot with the trained neural network as a production manager against the built-in Terran bot. The probabilistic strategy selects actions with probabilities equal to the outputs of the network while the greedy network always selects the action with the highest output, and random always picks a random action. The blind probabilistic network does not receive information about the opponent's material (inputs are set to 0.0). UAlbertaBot playing as Protoss with the scripted dragoon rush strategy wins 100\% of all games against the built-in Terran bot. 
  }\vspace{-0.2in}
\label{tab:results}
\end{table}

\begin{figure*}
  \includegraphics[width=\textwidth]{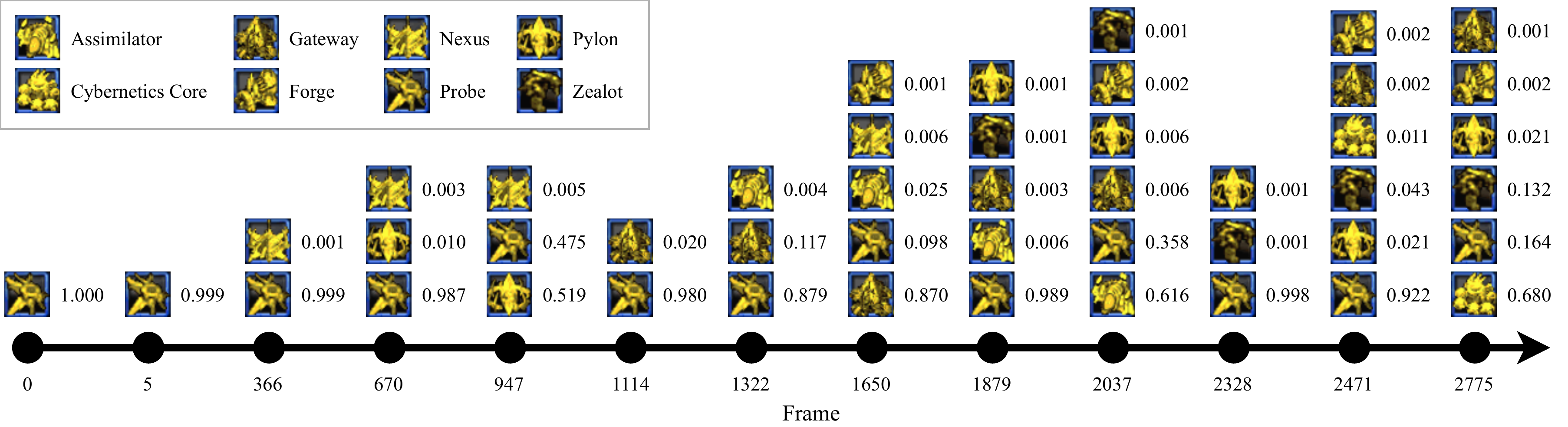} 
  \caption{The opening build order learned by the neural network when playing against the built-in Terran bot (the build order also depends on the enemy units observed). The number next to each build icon represents the probability of the build being produced next, and  points on the timescale indicate when the bot requests the network for the next build. In this example the network follows the greedy strategy, always picking the build with the highest probability.}\vspace{15pt}
  \label{fig:build_order}
\end{figure*}

\begin{table*}
\begin{center}
\begin{tabular}{ |l|ccccccc|cccc| } 
 \hline
 &
 \footnotesize{Probe} & 
 \footnotesize{Zealot} & \footnotesize{Dragoon} & \centering \pbox{20cm}{\footnotesize{$\,\,\,$Dark} \\ \footnotesize{templar}} & \footnotesize{Observer}
 & \footnotesize{Scout} & \footnotesize{Corsair} &
  \pbox{20cm}{\footnotesize{$\,\,\,\,\,\,\,\,\,\,$ Leg}\\
 \footnotesize{enhancements}} & 
 \pbox{20cm}{\footnotesize{$\,$Ground}\\
 \footnotesize{weapons}} & 
 \pbox{20cm}{\footnotesize{Ground}\\
 \footnotesize{$\,\,$armor}} &
 \pbox{20cm}{\footnotesize{Plasma}\\
 \footnotesize{shields}}\\
 Action selection &
\inlinegraphics{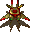} &
\inlinegraphics{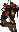} & \inlinegraphics{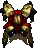} & \inlinegraphics{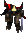} & \inlinegraphics{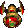} 
 & \inlinegraphics{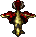} &
 \inlinegraphics{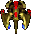} &
 \inlinegraphics{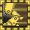} & 
\inlinegraphics{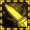} &
\inlinegraphics{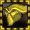} &
\inlinegraphics{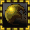}
 \\
 \hline
 \hline
 Probabilistic & 50.84 & 14.62 & 17.3 & 1.00 & 3.56 & 0.11 & 0.13 & 0.32 & 0.03 & 0.07 & 0.01 \\
 Greedy & 70.12 & 1.46 & 32.75 & 0.00 & 2.40 & 0.00 & 0.00 & 0.00 & 0.00 & 0.00 & 0.00\\
 \hline
 \end{tabular}
 \vspace{5pt}
\caption{The average number of different unit types produced by the two different action selection strategies against the built-in Terran bot. The results show that the greedy strategy executes a very one-sided unit production while the probabilistic strategy is more varied. }
\label{tab:builds-produced}\vspace{-15pt}
\end{center}
\end{table*}
Figure~\ref{fig:build_order} visualizes the learned opening strategy with greedy action selection. While the probabilistic strategy shows a better performance in general (Table~\ref{tab:results}), the strategy performed by the greedy action selection is easier to analyze because it is deterministic and has a one-sided unit production.
The learned build order shown in Figure~\ref{fig:build_order} is a \emph{One Gate Cybernetics Core} opening with no zealots before the cybernetics core. This opening was performed regularly against the built-in Terran bot, which does not vary much in its strategy. 
The opening is followed by a heavy production of dragoons and a few observers. A base expansion usually follows the first successful confrontation. Some losses of the greedy approach were caused by UAlbertaBot not being able to produce more buildings, possibly because there was no more space left in the main base. A few losses were also directly caused by some weird behavior in the late game, where the bot (ordered by the neural network) produces around 20 pylons directly after each other. Generally, the neural network expresses a behavior that often prolongs the game, as it prefers expanding bases when leading the game. This is something human players also tend to do, but since UAlbertaBot does not handle the late game very well, it is not a good strategy for this particular bot.

The behavior of the probabilistic strategy is more difficult to analyze, as it is stochastic. It usually follows the same opening as the greedy approach, with small variations, but then later in the game, it begins to mix its unit production between zealots, dragoons and dark templars. The timings of base expansions are very different from game to game as well as the use of upgrades. 

\section{Discussion}
This paper demonstrated that macromanagement tasks can be learned from replays using deep learning, and that the learned policy can be used to outperform the built-in bot in StarCraft. In this section, we discuss the short-comings of this approach and give suggestions for future research that could lead to strong StarCraft bots by extending this line of work. 

The built-in StarCraft bot is usually seen as a weak player compared to humans. It gives a sufficient amount of competition for new players but only until they begin to learn established opening strategies. A reasonable expectation would be that UAlbertaBot, using our trained network, would defeat the built-in bot almost every time. By analyzing the games played, it becomes apparent that the performance of UAlbertaBot decrease in the late game. It simply begins to make mistakes as it takes weird micromanagement decisions when it controls several bases and groups of units. The strategy learned by our network further enforces this faulty behavior, as it prefers base expansions and heavy unit production (very similar to skilled human players) over early and risky aggressions. The trained  network was also observed to make a few faulty decisions, but rarely and only in the very late game. The reason for these faults might be because some outputs are excluded, since UAlbertaBot does not handle these builds well.

Despite the presented approach not achieving a skill level on pair with humans, it should be fairly straightforward to extend it further with reinforcement learning. Supervised learning on replays can be applied to pre-train networks, ensuring that the initial exploration during reinforcement learning is sensible, which proved to be a critical step to surpass humans in the game Go \cite{silver2016mastering}. Reinforcement learning is especially promising for a modular-based bot as it could optimize the macromanagement policy to fit the fixed micromanagement policy. Additionally, learning a macromanagement policy to specifically beat other bots that are competing in a tournament is a promising future direction. 

This paper also introduces a new benchmark for machine learning, where the goal is to \emph{predict the next unit, building, technology or upgrade that is produced by a human player given a game state in StarCraft}. An interesting extension to the presented approach, which could potentially improve the results, could involve including positional information as features for the neural network. The features could be graphical and similar to the minimap in the game that gives an abstract overview of where units and buildings are located on the map. Regularization techniques such as dropout \cite{srivastava2014dropout} or L2 regularization \cite{nowlan1992simplifying} could perhaps reduce the error rate of deeper networks and ultimately improve the playing bot.

Finally, it would be interesting to apply our trained network to a more sophisticated StarCraft bot that is able to manage several bases well and can control advanced units such as spell casters and shuttles. This is currently among our future goals, and hopefully this bot will participate in the coming StarCraft competitions.

\section{Conclusion}
This paper presented an approach that learns from StarCraft replays to predict the next build produced by human players. 789,571 state-action pairs were extracted from 2,005 replays of highly skilled players. We trained a neural network with supervised learning on this dataset, with the best network achieving top-1 and top-3 error rates of 54.6\% and 22.9\%. To demonstrate the usefulness of this approach, the open source StarCraft bot UAlbertaBot was extended to use such a neural network as a production manager, thereby allowing the bot to produce builds based on the networks predictions.
Two action selection strategies were introduced: A greedy approach that always selects the action with the highest probability, and a probabilistic approach that selects actions corresponding to the probabilities of the network's softmax output. The probabilistic strategy proved to be the most successful and managed to achieve a win rate of 68\% against the games built-in Terran bot. Additionally, we demonstrated that the presented approach was able to play competitively against UAlbertaBot with a fixed rush strategy. Future research will show whether reinforcement learning can improve these results further, which could narrow the gap between humans and computers in StarCraft.

\end{document}